\documentclass[10pt]{article}

\usepackage[utf8]{inputenc}
\usepackage[]{titlesec}
\usepackage{booktabs,subcaption,amsfonts,dcolumn}
\usepackage{stfloats}
\usepackage{array}
\usepackage{diagbox}
\usepackage{algorithmic}
\usepackage[cmex10]{amsmath}
\usepackage[ruled,vlined,linesnumbered,noresetcount]{algorithm2e}

\usepackage[
backend=biber,
style=ieee,
maxbibnames=8,
maxcitenames=8,
]{biblatex}
\title{Structured Deep Neural Network Pruning via Matrix Pivoting}
\author{Ranko~Sredojevic$^{1,2}$,
        Shaoyi~Cheng$^2$,
        Lazar~Supic$^1$,
        Rawan~Naous$^1$,
        \\Vladimir~Stojanovic$^1$
}
\date{\small $^1$UC Berkeley $^2$numericcal\footnote{numericcal.com}\\\{rrs,sh\_cheng,lazar,rawansn,vlada\}@eecs.berkeley.edu}

\usepackage{graphicx}

\titleformat{\paragraph}
{\normalfont\normalsize\bfseries}{\theparagraph}{1em}{}
\titlespacing*{\paragraph}
{0pt}{3.25ex plus 1ex minus .2ex}{1.5ex plus .2ex}

\newtheorem{observation}{Observation}

\begin{document}

\maketitle

\begin{abstract}
Deep Neural Networks (DNNs) are the key to the state-of-the-art 
machine vision, sensor fusion and audio/video signal processing. 
Unfortunately, their computation complexity and tight
resource constraints on the Edge make them hard to 
leverage on mobile, embedded and IoT devices.
Due to great diversity of Edge devices, DNN designers have to take
into account 
the hardware platform and application requirements
during network training. 
In this work we introduce {\it pruning via matrix pivoting}
as a way to improve network pruning by compromising between
the design flexibility of architecture-oblivious
and performance efficiency of architecture-aware pruning, the two
dominant techniques for obtaining resounce-efficient DNNs. We also describe
local and global network optimization techniques for efficient
implementation of the resulting pruned networks. In combination, the proposed pruning
and implementation result in close to linear speed up with the reduction
of network coefficients during pruning.

\end{abstract}

\section*{Keywords}
C.3.d Real-time and embedded systems,
I.2.6.g Machine learning,
I.5.1.d Neural nets,
G.1.3.i Sparse, structured, and very large
systems,
I.3.1.a Graphics processors,
D.3.4.g Optimization

\section{Motivation}
\label{sec:motivation}

In order to build smart edge-based applications we must enable edge devices to leverage
state-of-the-art algorithms for data processing and information 
extraction. Many of those algorithms are based on DNNs \cite{deep_learning,energy_pruning}. 
The problem of efficient implementation of DNN architectures on edge
devices has received significant attention in recent
literature \cite{connections_then_weights, scalpel, structured_pruning, energy_pruning}.
However, the large number of different techniques reported
indicates that there is no convergence and it is unlikely that 
a one-size-fits-all technique will be found.
In fact, we will see examples showing that
appropriate techniques depend on the application, the target hardware
and the data set. Hence, we must navigate the trade-off between 
the network accuracy, size, energy and latency during hyperparameter
tuning phase of network training. The best course of action in this
situation is to discover, describe and characterize a large number
of potentially useful techniques for navigating these trade-offs 
when designing DNNs for edge applications.

Proposed solutions fall into three broad categories
\begin{itemize}
    \item {\bf Reduced-size network architectures} \cite{mobilenet,squeezenet} impose size and 
    structure constraints at the DNN level that are assumed to be appropriate for implementation on smaller
    devices.
    \item {\bf Layer pruning techniques} \cite{energy_pruning,connections_then_weights} try to reduce the number of 
    non-zero coefficients in layer weights, hoping for reduced number of operations as well as memory transfers.
    \item {\bf Scalar tricks} \cite{xnor,qnn} such as quantization or format conversion work by reducing the amount of storage or compute necessary for handling one scalar operation. 
\end{itemize}

This work introduces architecture-aware structured network pruning and 
implementation techniques optimized for low latency inference. Building on the previous work in architecture-aware pruning \cite{scalpel,energy_pruning}, we introduce structured transformations that bridge the gap between allowing the flexibility in significant coefficient locations at the network layer level and structured groupings characteristic of most hardware architectures. These techniques allow us to achieve close to linear speed up, with
reduction of layer coefficients, on modern embedded and mobile GPUs. Network training examples demonstrate marginal accuracy loss when imposing the proposed structural constraints during pruning. 

In latency optimization, fully-connected (FC) and late stage convolutional layers are the bottleneck.
While it is widely acknowledged that majority of computation in modern DNNs happens in early convolutional layers \cite{energy_pruning}, it is not widely understood that those layers are significantly easier to optimize through numerous opportunities
for latency hiding and data reuse \cite{efficient_mem_prog}. For example, in NVidia DriveNet running on Qualcomm Adreno 530 GPU 
early convolutional layers account for as much as $\approx$ 75\% of all operations, but only $\approx$ 45\% of inference latency. The last $\approx$ 3\% of operations are responsible for $\approx$ 25\% of the total DNN latency. Finally, while convolutional neural networks are widely used in literature for pruning studies, most networks (recurrent neural nets and multi-layer perceptrons) critical for day-to-day applications widely utilize the FC layers \cite{tpu_v2}. Hence, we will use the FC layers to demonstrate the proposed technique.

\subsection{Contributions}

The contributions of this work include:
\begin{itemize}
    \item {\bf Section \ref{sec:arch_insights}} A novel technique to avoid gather-scatter problem when handling sparse matrix
    vector multiplication arising from pruning of FC layers. The technique relies on a representation
    of structured sparse matrices in what we call a {\bf permutation-block-permutation (PBP)} form.
    \item {\bf Section \ref{sec:training}} Evidence that we can impose the PBP structure
    constraint in network training without significant negative impact on network accuracy. 
    We show examples of PBP pruning of networks for MNIST, CIFAR10. The technique does not
    lose any accuracy on MNIST and stays within 0.5\% of the original network performance on CIFAR10.
    \item {\bf Section \ref{sec:impl_single}} An efficient strategy for single-layer 
    (local) optimization of pruned FC layers by leveraging the PBP form.
    \item {\bf Section \ref{sec:impl_multi}} A set of compiler optimization techniques to
    leverage PBP form across multiple back-to-back FC layers.
    \item {\bf Section \ref{sec:results}} Experimental evidence that the proposed technique
    leads to significant improvement in inference execution speed when compared to the
    vendor-optimized dense and sparse linear algebra routines. Specifically, our approach
    leads to close to linear speedups with the reduction of matrix coefficients during pruning
    in all but fringe cases - small or highly sparse matrix, in which case it still
    outperforms other options.
\end{itemize}

\section{Architectural insights}
\label{sec:arch_insights}

In general, pruning of DNNs will result in sparse layer matrices \cite{connections_then_weights}.
This allows programmers to leverage sparse matrix-vector multiplication when optimizing 
DNNs for inference. Unfortunately, generic sparse matrix libraries are tuned for fill-in
levels below 1\%, well below the fill-in of more than 5\% we see in DNN
pruning literature \cite{energy_pruning,connections_then_weights,structured_pruning,scalpel}. Hence,
simply relying on generic sparse matrix-vector routines leaves significant performance
on the table. We can observe this in recent works where speed increase due to sparsity
lags significantly behind the reduction in coefficients \cite{scalpel,structured_pruning}. 

The reason for this inefficiency can be found if we observe the memory bandwidth utilization
in dense and sparse matrix-vector multiplication. Typically, dense matrix-vector multiplication achieves
excellent load and store bandwidth utilization. However, sparse matrix-vector multiply
typically achieves good load bandwidth utilization, but very poor store utilization. 
Measurements of the load and store bandwidth utilization for 
different implementations can be found in Section \ref{sec:res_util}.

As a result, a variety of {\it structured pruning} \cite{structured_pruning,scalpel} techniques were proposed where full
blocks, columns or rows of layer weight matrices are removed. These techniques enables the
use of dense linear algebra kernels despite the reduction in the size of layer weight matrices.
However, this efficiency comes with a harsh constraint that we must remove a whole row/column if
we want to remove a single element. 

In this work, we propose a novel {\it structured transformations pruning} approach that enables low latency implementation
by constraining the sparse representation of pruned layers during computation while allowing for high flexibility of non-zero coefficient positioning in the network representation. 
The key idea is that network is pruned in such a way that there exists a convenient tensor transformation that turns the pruned network representation into a desirable computational structure.
If we know this map at network implementation time we could account for it in our
parallelization and vectorization strategies to achieve high runtime efficiency just like
in structured pruning. At the same time, the tensor transformation should enable more
flexible positioning of non-zero layer weights just like in unstructured pruning.

\begin{figure*}[!t]
    \centering
    \includegraphics[width=0.95\textwidth]{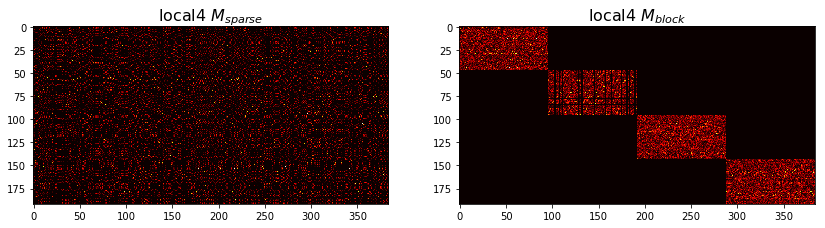}
    \caption{An example of $(M_{sparse}, M_{block})$ pair}
    \label{fig:sparse_block}
\end{figure*}

One way to do this is to apply a standard technique from numerical linear algebra
and high-performance real-time optimization \cite{cvxgen}: compile-time matrix pivoting.
Namely, for any fully connected layer matrix, we will perform the pruning to obtain $M_{sparse}$ that can be represented in what we call the PBP form
\begin{equation}
\label{eqn:permutation-block}
    M_{sparse} = P_{row} M_{block} P_{col}
\end{equation}
\noindent where $P_{row/col}$ are row and column permutation matrices and $M_{block}$ is the computationally desirable matrix form tailored to a given underlying architecture. If such representation results in good network accuracy, we will have achieved our goal by encapsulating all computation and memory access irregularities into permutation actions. Furthermore,
$P_{row/col}$ will be static, i.e. these permutations depend on the sparsity pattern of the pruned matrix, only. Specifically, they do not depend on any runtime data. An example of $(M_{sparse}, M_{block})$
pair of matrices is shown in Figure~\ref{fig:sparse_block}. As long as the overhead of the permutations is not larger than the savings due to structured computation, the approach should lead to speed-up.

Once we have the PBP form of an FC layer, we can leverage associativity of matrix multiplication
to isolate irregular memory access patterns in implementation as follows:
\begin{equation}
\label{eqn:pbp_implementation}
a_{out} = M_{sparse} a_{in} = (P_{row} M_{block} P_{col}) a_{in} = P_{row} (M_{block} (P_{col} a_{in}))
\end{equation}

In general, there is no reason to believe that such representation will be possible
after unconstrained (proxy-based) pruning. Hence, we need to nudge the network to learn the sparse structure in such a way as to
guarantee that the transformation in Eqn. \ref{eqn:permutation-block} is possible.

\section{PBP pruning}
\label{sec:training}

In the previous section we explained the intuition behind the PBP
representation of pruned FC layers. Now we must evaluate the impact
of imposing the PBP constraint on network accuracy.

There are different ways to obtain PBP form during network training. We will describe two techniques
and how they perform on TensorFlow tutorial examples for MNIST and CIFAR10 data sets. Coefficients
for all example networks from this section, including the permutations needed for transitioning 
from $M_{sparse}$ to $M_{block}$ form, will be available online.

\subsection{Feed-forward PBP pruning}

A particularly simple way to obtain a PBP form of an FC layer is to just choose $P_{row/col}$ before training and enforce it during training. This technique relies on coefficient redundancy in the network and, somewhat surprisingly, works quite well on some networks.
When the resulting accuracy is acceptable, this approach is particularly convenient as it opens up a number of opportunities for
cross-layer optimization, as we will see in Section~\ref{sec:impl_multi}.

The baseline accuracy of TensorFlow Deep MNIST tutorial is 99.2\%, which is achieved within 20,000
minibatch steps, with batch size $B=128$. Picking two random permutations for rows and columns
and setting the block structure for 6.125\% and 50.0\% fill-in on the {\it fc1 and fc2}, respectively,
results in accuracy between 99.1\% and 99.3\%. It should be noted that picking identity permutations
results in loss of accuracy over 10\%. However, picking random permutation results in no accuracy
loss with very high probability. We run hundreds of training sessions with randomly chosen permutations and
have not seen any of them lose accuracy.

As the second example, we use the TensorFlow CIFAR10 convolutional network tutorial. The baseline
network accuracy, achieved within 125,000 minibatch steps, is (87.0 $\pm$ 0.1)\%. The feed-forward
approach to finding a PBP form with 12.5\% fill-in, loses $\approx$ 1.6\% accuracy. In such cases, to further optimize the trade-off between the fill-in and network accuracy we must search for more efficient ways to impose the PBP structure during network training.

\subsection{Feed-back PBP pruning}

An alternative to the feed-forward PBP pruning is the feed-back pruning, where we 
first train the network without pruning in order to determine where large, presumably more important,
connections form in each layer \cite{connections_then_weights}. We can use that information to pick permutations that would let us keep
as many of such connections as possible. This problem is similar to graph bisection, and it cannot be
solved exactly. Hence, we resort to a heuristic search.

We start by training the network without any pruning. At this point, for each layer we intend to prune, we look at the absolute values of weights in the layer matrix and attempt to pick $P_{row/col}$ that would maximize the sum of absolute values of coefficients in the upper left and the lower right sub-block of the layer matrix. This is done by greedy optimization starting from
randomly chosen permutations. The coefficients that end up in the sub-blocks off the main diagonal 
are simply deleted and the training continues. Each time we repeat this process on a sub-block we
delete 50\% of coefficients in the block.

In our tests we performed the bisection of all blocks (the number of blocks doubles after each
bisection) three times on {\it local3} and two times on {\it local4}
layers of the TensorFlow CIFAR10 tutorial network, resulting in 12.5\% and 25.0\% fill-in factors. 
After 200,000 minibatch steps (not including the original training) the network converges to 86.7\% 
accuracy. Figure~\ref{fig:sparse_block} shows the resulting {\it local4}
layer matrix in both the sparse and blocked form. 

\section{Implementation}
\label{sec:impl}

In the previous two sections, we explained the intuition behind the PBP form and 
demonstrated a few possible ways to obtain a network that satisfies such a constraint. In this section we return to
Equation~\ref{eqn:pbp_implementation} and describe in detail a high-performance implementation for mobile
class GPUs.
We will handle local (single-layer) and global (multi-layer) optimization separately.

\subsection{Single-layer optimization}
\label{sec:impl_single}

Given the different computational nature of permutation and block-matrix/vector multiplication we will handle them
separately. 

\begin{figure*}[!t]
    \centering
    \begin{subfigure}[t]{0.28\textwidth}
        \centering
        \includegraphics[width=0.68\textwidth]{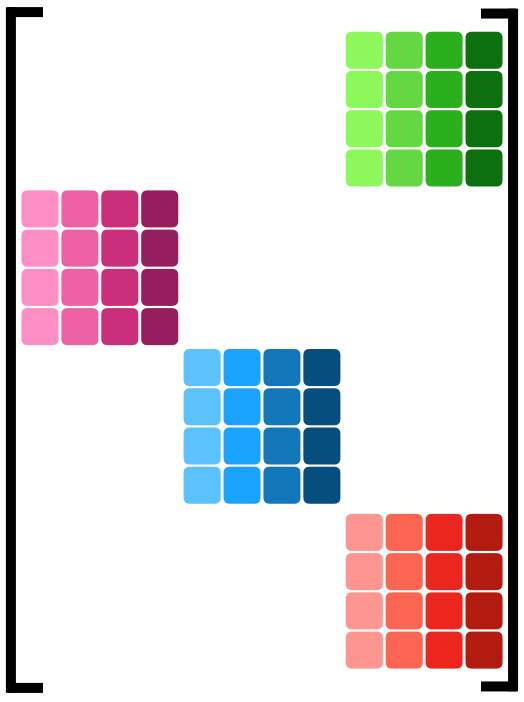}
        \caption{Logical view}
        \label{fig:logical_layout}
    \end{subfigure}
    \hfill
    \begin{subfigure}[t]{0.35\textwidth}
        \centering
        \includegraphics[width=\textwidth]{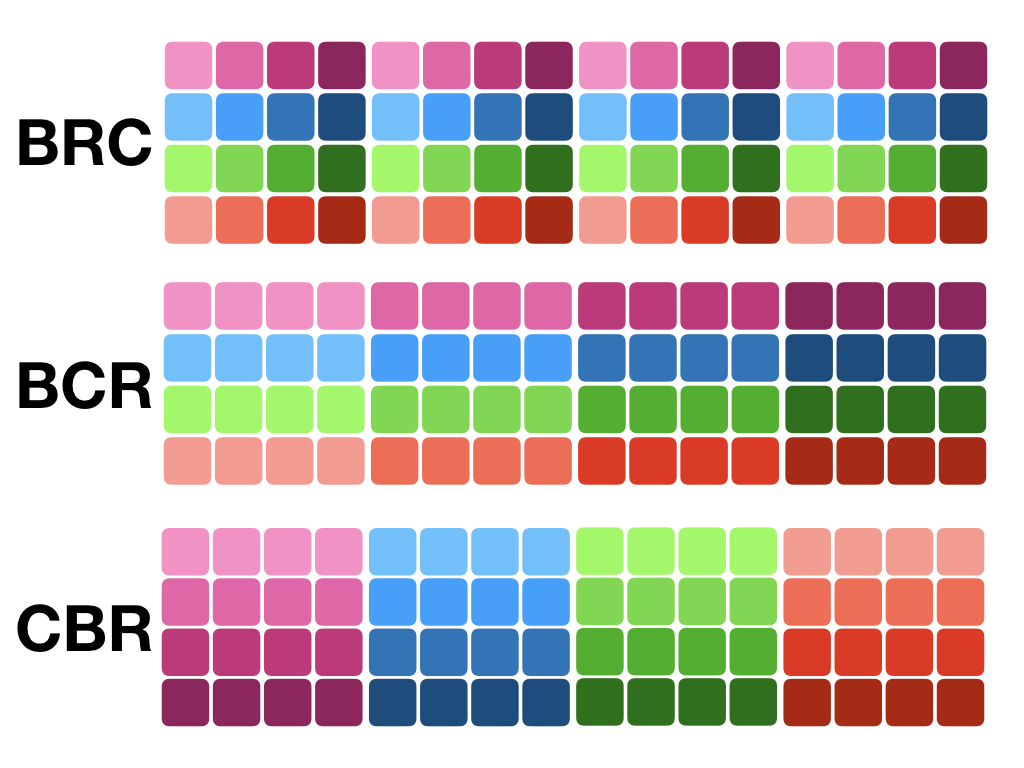}
        \caption{Physical layouts}
        \label{fig:physical_layout}
    \end{subfigure}
    \begin{subfigure}[t]{0.35\textwidth}
        \centering
        \includegraphics[width=\textwidth]{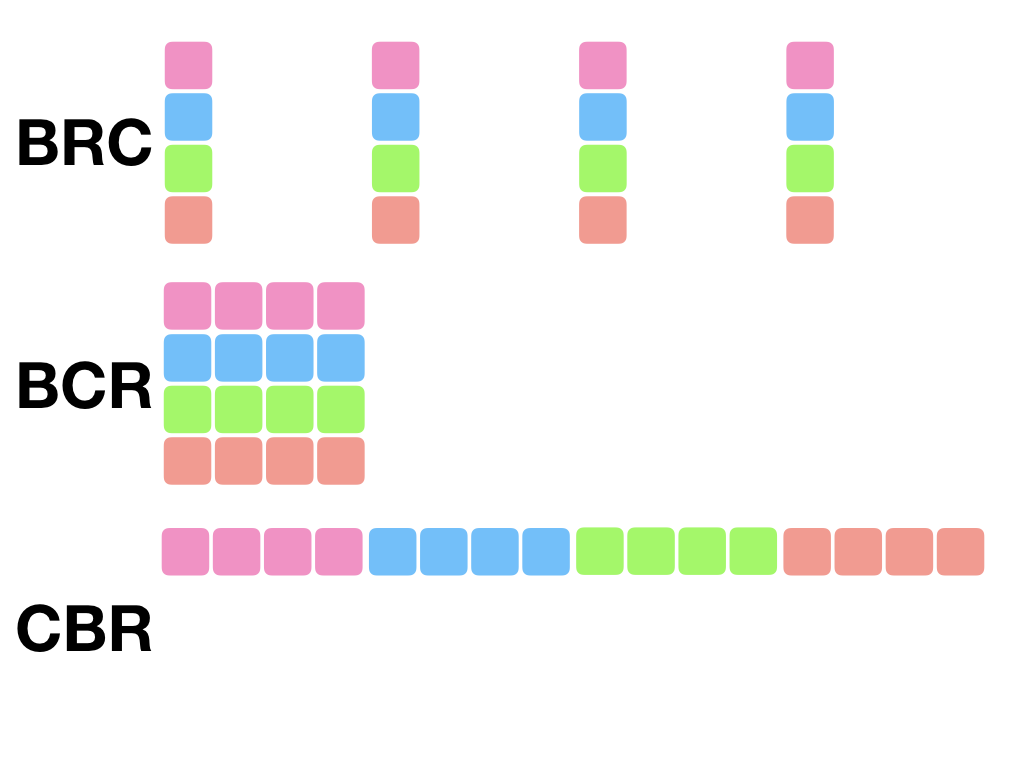}
        \caption{Access pattern}
        \label{fig:access_pattern}
    \end{subfigure}
    \caption{Different logical to physical memory mapping of $M_{block}$}
    \label{fig:memory_layouts}
\end{figure*}

\subsubsection{Block-matrix multiplication}
\label{sec:blk_mul}

Let us consider a sparse block-matrix shown in Figure~\ref{fig:logical_layout}. We assume that each
colored block is potentially a dense submatrix, while the white areas represent zero coefficients. For now,
we assume that blocks are identical squares that occupy non-overlapping sets of rows. We will comment on handling of different block and matrix shapes in Section~\ref{sec:irregular_matrices}. We order the blocks arbitrarily, for example in {\it purple-blue-green-orange} order,  as we map them to physical memory of
the GPU. 

With Figure~\ref{fig:logical_layout} in mind, our goal is to efficiently implement 
the $M_{block}$ multiplication from Equation~\ref{eqn:pbp_implementation}.
We approach the GPU implementation with the following model in mind:
\begin{itemize}
    \item We will launch {\bf one thread group per block}. 
    \item Within a thread group will have {\bf one thread per output coordinate}, i.e. each number in the
          output (activation) vector will be calculated by one thread working along the row
          of a block (sub-matrix).
\end{itemize}

With these guidelines in mind, it is simple to implement efficient GPU thread kernels (e.g. OpenCL, CUDA, Vulkan).
As usual, a number of implementation tricks like loop-unrolling and scalar promotion \cite{efficient_mem_prog} 
are useful in tuning performance. 
The biggest improvement, however, comes from aligning and coalescing memory accesses. 

To optimize memory access we consider three different memory layouts, shown in Figure~\ref{fig:physical_layout}. Later, in the Section~\ref{sec:impl} we will see that the correct choice of the memory layout depends on the target computer architecture. 
Interestingly, most results to date typically argue for one specific memory layout overlooking the strong interaction between the computer architecture and performance optimization.
Capturing this strong interaction effectively is the biggest challenge in bringing high-quality, easy-to-use DNN technology to the heterogeneous hardware platforms on the Edge.

In the image we represent memory as an array of cells with consecutive addresses moving left-to-right/top-to-bottom.
We name the layouts after the tensor axis layout order using rightmost axis changing fastest rule. We indicate the 
correspondence between the logical layout in Figure~\ref{fig:logical_layout} and physical layouts in
Figure~\ref{fig:physical_layout} using color shades:
\begin{enumerate}
    \item {\bf Block-Row-Column (BRC)} layout, or {\bf row-major within block}, fixes the block and lays out 
          each block row consecutively in memory. Hence, column coordinate within each row changes with each
          consecutive memory location. 
    \item {\bf Block-Column-Row (BCR)} layout, or {\bf column-major within block}, fixes the block and lays out
          each block column consecutively in memory. 
    \item {\bf Column-Block-Row (CBR)} layout tries to maximize memory coalescing by fixing the column and then laying
          out that column position from all blocks consecutively in memory. 
\end{enumerate}

Finally, Figure~\ref{fig:access_pattern} demonstrates the memory reads across all threads
on the first computation cycle after launching a set of thread groups. We should keep in mind that we parallelize
on output, with each thread handling one output activation coordinate. Hence,
the first column of each block will be fetched first after launching the grid. 

As expected, the preferred memory layout
depends on the architectural features of the underlying GPU device. We now briefly review architectural features
of different device families to understand access pattern influence on performance.

\subsubsection{Implementation tradeoffs on NVidia Maxwell GPU}

NVidia GPUs (and similar devices) use Single-Instruction-Multiple-Threads 
(SIMT) model. Threads are grouped into {\it warps} and warps into {\it blocks}. While each thread has its own state, including the Program Counter, maximum device utilization is achieved if all threads within a warp execute the same instruction on the same cycle.
Furthermore, we aim to align memory accesses across threads in a group to minimize the number of bus transactions
needed to fetch and store data.

Typically, at the start of kernel execution all data is in the global GPU memory, uploaded by the host system.
In this case all the blocks are in global memory regardless of the layout scheme. In our experiments, due
to low compute per fetch in FC layer processing, prefetching from global memory through combined efforts of the thread group was not effective or needed.

The latency of global memory fetch can be hidden well most of the time through instruction level parallelism 
within a scalar thread.
There are a number of independent operations to be performed within each thread, essentially implementing a dot
product between the assigned block row and a subsection of the vector being multiplied. We also aid
the GPU compiler by summing consecutive fetches to multiple partial register accumulators and reducing them
at the end of thread's execution. 

In this situation where all warp threads share memory bandwidth and run in lockstep the preferred memory layout
is CBR as it provides the best memory coalescing. As we will see in Section~\ref{sec:results} this is confirmed
in our experiments. In practice, the BCR comes close second. When the amount of data fetched is extremely small,
e.g. small matrices with low fill-in, BCR is a little slower as it wastes bandwidth by forcing multiple fetches
due to gaps in memory accessed on the same cycle, Figure~\ref{fig:access_pattern}. However, for higher fill-in
levels there is sufficient data in every strip of memory in Figure~\ref{fig:access_pattern} for BCR to
effectively use the memory bandwidth and this difference disappears. BRC layout, Figure~\ref{fig:physical_layout}, uses memory bandwidth very inefficiently, Figure~\ref{fig:access_pattern}, 
and that shows both in load efficiency and in computation latency measurements. 

\subsubsection{Implementation tradeoffs on ARM Mali GPU}

The ARM Mali GPU, in particular the Midgard architecture, has a different parallelization model than Nvidia devices. 
Even though it is also a massively multithreaded computation platform, where large number of threads are needed to keep the device
busy and hide memory access latency, it does not use the SIMT paradigm. Instead of bundling multiple threads together and pushing them through the array of ALUs, Mali executes each thread independently. 

Meanwhile, cross-thread memory coalescing, an important factor in achieving good performance in Nvidia devices, 
 does not result in better speed here. BRC layout, whose performance lags both BCR and CBR layout when Nvidia GPU is used, 
 achieves the fastest speed in Mali. With BRC, every individual thread (instead a warp of threads) fetches a contiguous chunk of data, which is more compatible with Mali's execution model. Section~\ref{sec:results} will present the achieved performance of this layout. 
 


\subsubsection{Handling permutations}
\label{sec:permutations}

In Section~\ref{sec:arch_insights} we described our plan to isolate all memory access irregularities into 
a separate processing step. This allowed us to pack $M_{block}$ into a set of dense blocks and produce an efficient
implementation described in the previous section. Now we must handle the irregular computation pattern, encapsulated in the row/column permutation matrices.

Unfortunately, there is no invariant or reasonable assumption we can introduce around permutations. We currently
do not have a way to impose constraints on permutations during PBP pruning (network training) without sacrificing too much
accuracy. Hence, we must handle $P_{row}$ and $P_{column}$ in a completely general way.

We represent permutations in global GPU memory as index arrays. For example, the index array of {\it "cab"} relative
to {\it "abc"} is $[2,0,1]$. Incoming vector, produced by the previous network layer, is also assumed to be stored in 
the global GPU memory. 

Looking at Figure~\ref{fig:logical_layout} we observe that each block only acts on a sub-slice of the incoming
vector $v_{in}$, as all weights that share row but not column coordinates with a particular block are zero. Hence, each
block only needs access to a subset of the input vector coordinates. Moreover, this subset is shared across the
block, meaning that it can be pre-fetched in parallel by all threads in the tread group working together.

Similarly, each block only produces a subset of the outgoing vector $v_{out}$ coordinates. As explained previously,
each thread in a thread block will process one row of the block-matrix. Hence, we could store values produced
by the block in thread group (shared) memory and update the global memory after all threads finish. 

Pseudo-code for an implementation of a GPU kernel along these lines is shown in Listing~\ref{list:two_permute}.

\begin{algorithm}[H]

\SetKwInOut{Input}{input}\SetKwInOut{Output}{output}

\SetAlgorithmName{Listing}

\SetAlgoLined

\Input{activation vector $a_{in}$, matrix $M_{block}$, index arrays $P_{row}$/$P_{col}$}
\Output{output activation vector $a_{out}$}
 identify the thread's global coordinate (Block, Row)\;
 identify the BlockColumns set\;
 identify the BlockRows set\;
 group fetch $a_{local}[] \leftarrow a_{in}[P_{col}[\text{BlockColumns}]]$\;
 $synchronize\_threads()$\;
 \For{c $\leftarrow$ 0 \KwTo BlockColumns}{
     $acc[Row] \leftarrow acc[Row] + M_{block}[Block][Row][c] a_{local}[c]$\;
 }
 $synchronize\_threads()$\;
 group store $acc[] \rightarrow a_{out}[P_{row}[\text{BlockRows}]]$
 \caption{PBP kernel with separate row/column permutation}
 \label{list:two_permute}
\end{algorithm}

Starting from this baseline implementation we will proceed to handle more general cases and 
optimize performance.

\subsubsection{Irregular matrix shapes}
\label{sec:irregular_matrices}

In the previous section we assumed that all blocks have the same size and that no two blocks share
any of their row coordinates within the block. It is simple to extend the described technique to a more
general setting. It is easy to see that blocks can have different sizes if we pass information about block sizing
to the kernel and use the $Block$ coordinate in Listing~\ref{list:two_permute} to
lookup the block the current thread is processing. If any two blocks share row coordinates we can have an additional reduction step after each block is processed as described previously.

\subsection{Cross-layer optimization}
\label{sec:impl_multi}

Though Listing~\ref{list:two_permute} can handle general permutations, they introduce overhead, as reported in Section~\ref{sec:perm_overhead}.
First, we are forced to perform the input fetch into $a_{local}$ that is not necessarily coalesced since
there is no constraint on $P_{col}$. Then we must synchronize threads after the group fetch. Similarly, we
must synchronize before storing all the accumulators from a thread group according to $P_{row}$ which, again,
does not guarantee a coalesced store.

Luckily, we can simplify and optimize the implementation by leveraging the algebraic properties of permutations
and their interaction with domain specific operators. To do this we must consider cross-layer interactions. First, we can remove the need for the $P_{row}$ in most situations.

\subsubsection{Cross-layer permutation fusion}
\label{sec:xlayer_perm}

Consider two consecutive PBP fully connected layers, represented with pruned matrices $A_{sparse}$ and
$B_{sparse}$. They would, typically, be connected through a non-linear, point-wise operator such as 
reLU. To see how we can leverage this situation we need two simple facts.

\begin{observation}
\label{obs:commute}
Permutations commute with point-wise operators such as reLU. In other words, whether we first permute
a vector and then apply a function to each element of the resulting vector has the same outcome as 
first applying the function and then permuting the result. Furthermore, permutations commute with 
softmax and similar operators.
\end{observation}

\begin{observation}
\label{obs:permute_comp}
The product of two permutation matrices, i.e. the composition of two permutations, is again a permutation.
\end{observation}

With this in mind we return to our case of back-to-back PBP FC layers. We can represent this
situation with the following pipeline section

\begin{align*}
    &\cdot \leftarrow A_{sparse} \leftarrow reLU \leftarrow B_{sparse} \leftarrow \cdot &\\
    &\cdot \leftarrow P^A_{row} M^A_{block} P^A_{col} \leftarrow reLU \leftarrow P^B_{row} M^B_{block} P^B_{col} \leftarrow \cdot & \text{using Equation~\ref{eqn:permutation-block}}\\
    &\cdot \leftarrow P^A_{row} M^A_{block} (P^A_{col} P^B_{row}) \leftarrow reLU \leftarrow M^B_{block} P^B_{col} \leftarrow \cdot & \text{using Observation~\ref{obs:commute}}\\
    &\cdot \leftarrow P^A_{row} M^A_{block} P^{AB}_{col/row} \leftarrow reLU \leftarrow M^B_{block} P^B_{col} \leftarrow \cdot & \text{using Observation~\ref{obs:permute_comp}}
\end{align*}

In other words, the row permutation $P_{row}$ implemented in lines 8 and 9 in Listing~\ref{list:two_permute} can 
be moved into the next layer and fused into the colum permutation $P_{col}$ implemented in lines 4 and 5. 
Furthermore, since PBP form relies on statically known permutations, the product $P^{AB}_{col/row}$ can be
determined at network implementation time,
by an optimizing domain specific compiler.
Hence, we do not need to handle the output permutation whenever two PBP sparse layers are back-to-back in
a DNN, which allows us to remove the processing in lines 8 and 9 in such cases.

\subsubsection{Output permutation elimination}

In classification tasks we can use a similar optimization at the output of the network. For example, extending
the previous example with an output softmax we can write
\begin{align*}
    & \leftarrow softMax \leftarrow P^A_{row} M^A_{block} P^{AB}_{col/row} \leftarrow \cdot & \\
    P^A_{row} &\leftarrow softMax \leftarrow M^A_{block} P^{AB}_{col/row} \leftarrow \cdot & \text{using Observation 1}
\end{align*}

At this point we can simply drop the final permutation $P^A_{row}$ since it simply
relabels the coordinates of the output
vector. The host application can take it into account when interpreting the result.

\subsubsection{Cross-layer permutation elimination}

Finally, in some situations we can completely eliminate all inference-time (run-time) overhead of permutations
between two PBP layers. To do this, we must check whether the PBP training can be adjusted to force
$P^{AB}_{col/row} = I$. In other words, we need to train the network in such a way to have $(P^A_{col})^T = P^B_{row}$.

\section{Results}
\label{sec:results}

To test the effectiveness of the proposed approach we have implemented the kernel described in
Listing~\ref{list:two_permute} with the cross-layer permutation fusion described in 
Section~\ref{sec:xlayer_perm}. All results are measured for square matrices with square, identically
sized blocks. For example, the case for $size = 512$ and $fill-in = 6.25$ was measured with 
$512 \times 512$ matrix with $16$ identical $32\times 32$ blocks. 

\subsection{Execution speed}

\begin{figure*}[!t]
    \centering
    \includegraphics[width=0.95\textwidth]{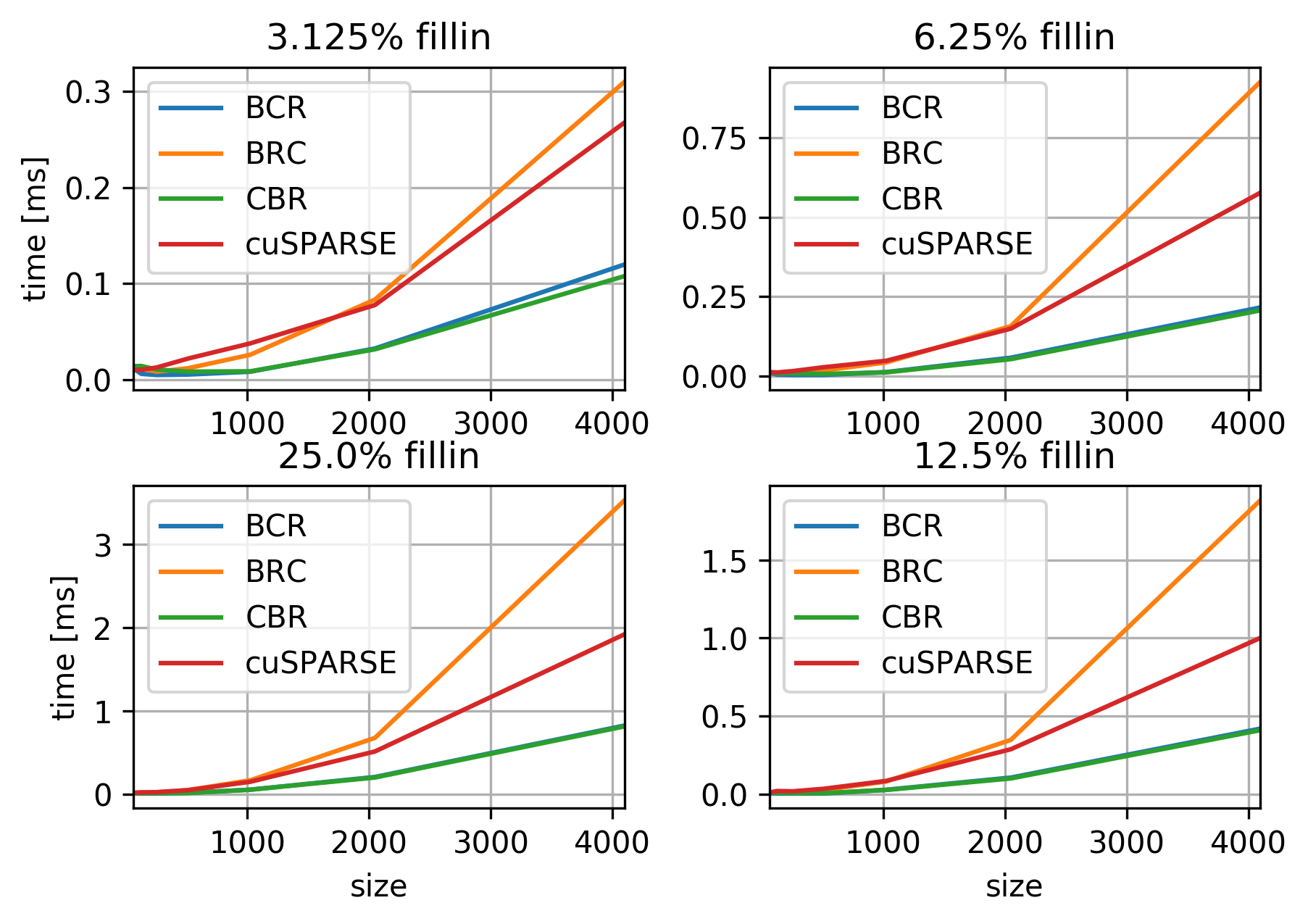}
    \caption{Average compute latency over 2,500 calls on Jetson TX1}
    \label{fig:latency_jetson}
\end{figure*}

The execution latency as a function of the matrix size and fill-in level on NVidia Jetson TX1 SoC with
Maxwell class GPU is given in Figure~\ref{fig:latency_jetson}. Results compare implementations with
the three memory layouts, CBR, BRC and BCR, against the NVidia cuSPARSE library, a generic sparse
linear algebra library typically used to implement pruned FC layers. The measurements support
the analysis in Section~\ref{sec:impl_single}.

Speedup of the proposed method with CBR layout, defined as the ratio between the latency 
of generic dense (cuBLAS) and sparse (cuSPARSE) matrix-vector multiplication to
the latency of the proposed implementation,
on NVidia Jetson TX1 is given in Table~\ref{tab:speedup_jetson}. On Mali GPU, speedup with BRC layout over the OpenCL dense matrix-vector
multiplication in ARM Compute Library is given in Table~\ref{tab:speedup_arm}. It is worth noting that this result for Mali T880 showed a great amount of variability. This is likely due to the fact that we have used an actual
cellphone for the experiment. Features ensuring pleasant user experience and prolonging battery life, e.g. aggressive thermal throttling and dynamic voltage frequency scaling, 
all affect the final measurements in non-deterministic ways.


\begin{table*}[!h]
    \centering
    \begin{subtable}{.95\textwidth}
    \centering
    \renewcommand{\arraystretch}{1.2}
    \begin{tabular}{c|c|c|c|c|c|c|c|c}
        \diagbox[width=5em]{size}{fill-in}
        & \multicolumn{2}{c|}{3.125\%} &
          \multicolumn{2}{c|}{6.25\%}  &
          \multicolumn{2}{c|}{12.5\%}  &
          \multicolumn{2}{c}{25.0\%} \\
    \hline
        64 & 2.00 & 0.86 & 2.37 & 1.22 & 4.13 & 1.56 & 3.36 & 2.04\\
        128 & 6.83 & 1.73 & 7.44 & 2.36 & {\bf 7.87} & 4.14 & {\bf 5.86} & 2.79\\
        256 & 22.61 & 2.71 & {\bf 28.35} & 4.22 & {\bf 22.55} & 3.80 & {\bf 16.15} & 3.55\\
        512 & 13.02 & 4.29 & {\bf 16.37} & 7.00 & {\bf 12.19} & 6.48 & {\bf 4.98} & 3.53\\
        1024 & 26.02 & 4.70 & {\bf 17.56} & 4.02 & {\bf 7.58} & 3.09 & {\bf 4.01} & 2.83\\
        2048 & 26.72 & 2.47 & {\bf 15.12} & 2.79 & {\bf 8.10} & 2.84 & {\bf 4.10} & 2.57\\
        4096 & {\bf 34.40} & 2.49 & {\bf 17.65} & 2.79 & {\bf 8.82} & 2.44 & {\bf 4.46} & 2.36\\
    \hline
    speedup vs & cuBL & cuSP & cuBL & cuSP & cuBL & cuSP & cuBL & cuSP
    \end{tabular}
    \caption{Speedup of PBP vs cuBLAS/cuSPARSE on Jetson TX1\newline}
    
    \label{tab:speedup_jetson}
    \end{subtable}
    \\
    \begin{subtable}{.75\textwidth}
    \centering
    \renewcommand{\arraystretch}{1.2}
    \begin{tabular}{c|c|c|c|c}
        \diagbox[width=5em]{size}{fill-in}
        & \multicolumn{1}{c|}{3.125\%} &
          \multicolumn{1}{c|}{6.25\%}  &
          \multicolumn{1}{c|}{12.5\%}  &
          \multicolumn{1}{c}{25.0\%} \\
    \hline
        64 & 3.90 & 3.64 & 3.51 & 2.98 \\
        128 & 5.71 & 5.44 & 4.74 & 4.28\\
        256 & 9.02 & 8.89 & 7.04 & 4.68 \\
        512 & 10.92 & 7.81 & 4.93 & 2.24 \\
        1024 & 9.69 & 5.33 & 2.36 & 2.09 \\
        2048 & 7.67 & 5.81 & 3.15 & 2.30 \\
        4096 & 14.09 & 9.26 & 6.97 & 3.85 \\
    \end{tabular}
    \caption{Arm Mali T880 on Huawei Honor 8}
    \label{tab:speedup_arm}
    \end{subtable}
    \caption{Speedup of the PBP gemv() over vendor libraries}
    \label{tab:speedup_jetson_arm}
\end{table*}

\subsubsection{Permutation overhead}
\label{sec:perm_overhead}
The overhead of performing permutation, as shown in lines 4 and 5 of
Listing~\ref{list:two_permute} is negligible for large blocks and high fill-in
factors. However, when fill-in factor is below 12.5\% and the block is
smaller than roughly $128\times 128$ on NVidia TX1, the time to perform one permutation
is $\approx$ 50\% of the kernel run time. The high overhead of permutation for small matrices and low fill-in factors is noticeable
in Table~\ref{tab:speedup_jetson}. Specifically, bold numbers
emphasize the cases where our implementation achieves linear and better
than linear speedup with the reduction of matrix coefficients. It is
obvious that there are no bold numbers along the edges and in the 
upper left corner of the table, exactly where the measured permutation
overhead was high. Most importantly, for fill-in factors of 6.25\% - 12.5\% and sizes larger than 128-256, which are typical values for the sparsest FC layers, permutation overhead is negligible.



\subsection{Resource utilization}
\label{sec:res_util}

In Section~\ref{sec:arch_insights} we conjectured that performance of a sparse matrix-vector multiplication
can be improved by constraining and encapsulating irregular memory access patterns in permutation operators.
We measured resource utilization on NVidia Jetson TX1 hardware using NVidia profiler, {\it nvprof}. 
Table~\ref{tab:resources} shows that PBP form, with Section~\ref{sec:xlayer_perm} optimizations, achieves
load and store efficiency on par with dense matrix multiply, significantly outperforming generic sparse
matrix-vector multiplication. 

\begin{table*}
\centering
    \renewcommand{\arraystretch}{1.2}
    \begin{tabular}{r|c|c|c}
    \diagbox[width=6em]{kernel}{metric} & gld\_efficiency & gst\_efficiency & occupancy \\
    \hline
    gemv2N & 100.0 & 100.0 & 54\\
    csrMv & 99.32 & 12.5 & 97\\
    brc & 12.52 & 100.0 & 52\\
    bcr & 98.1 & 100.0 & 42\\
    cbr & 98.7 & 100.0 & 41\\
    \end{tabular}
    \caption{GPU resource utilization \% as reported by NVidia nvprof on gemv()}
    \label{tab:resources}
\end{table*}

\section{Conclusion}
Deep Learning has the potential to become widely utilized in many applications on the Edge. 
However, for widespread use of DNNs in embedded and mobile software
to take place, we must ensure that design, optimization, integration and management of DNNs
alongside the standard business logic is straightforward and efficient. 
As a community we have a long way to go before we reach the moment when application
developers will not think twice before reaching for a DNN to solve their 
problem or improve a feature of their software.

In this work we developed {\it structured sparsity under transformation} for the Edge Machine Learning 
toolbox. We demonstrated how static matrix pivoting can be used to induce 
computationally friendly PBP structure in sparse DNN layers and demonstrated architecture-specific
optimization of PBP layers.

Going forward we expect that the next generation of DNN frameworks will appear, specifically
optimized for Edge deployment. Unlike the current solutions, the new generation will need to be organized as domain specific
compilers, not simple interpreters. Such transition will enable the use of domain-specific cross-layer optimization opportunities and fine-grained architecture-specific DNN tuning like the ones described in this paper. The outcome will be more efficient and productive DNN design iteration enabling faster innovation
and deployment on the Edge.

\printbibliography

\end{document}